\documentclass[conference]{IEEEtran}
\IEEEoverridecommandlockouts
\usepackage{cite}
\usepackage{amsmath,amssymb,amsfonts}
\usepackage{algorithmic}
\usepackage{graphicx}
\usepackage{textcomp}
\usepackage{xcolor}
\usepackage{comment,float}
\usepackage{subfig}
\usepackage{graphicx}
\usepackage{pgf}
\usepackage{amsmath}
\usepackage{times}
\usepackage{epsfig}
\usepackage{graphicx}
\usepackage{amsmath}
\usepackage{amssymb}
\usepackage{array}
\usepackage{hyperref}
\usepackage{arydshln}
\newcolumntype{M}[1]{>{\centering\arraybackslash}m{#1}}
\usepackage{multirow}

\usepackage[linesnumbered,vlined,algo2e]{algorithm2e}
\SetKwInput{KwInput}{Input}                
\SetKwInput{KwOutput}{Output}
\SetKw{KwInit}{Initilialise}

\usepackage{algorithm}
\def\BibTeX{{\rm B\kern-.05em{\sc i\kern-.025em b}\kern-.08em
		T\kern-.1667em\lower.7ex\hbox{E}\kern-.125emX}}
\begin{document}
	
	\title{Stochastic sparse adversarial attacks\\
	}
	

	\author{
		
		\IEEEauthorblockN{Manon Césaire}
		\IEEEauthorblockA{\textit{
				Institute of Research } \\
			\textit{and Technology SystemX} \\
			Palaiseau, France \\
			manon.cesaire@irt-systemx.fr}
		\and
		\IEEEauthorblockN{Lucas Schott}
		\IEEEauthorblockA{\textit{
				Institute of Research } \\
			\textit{and Technology SystemX} \\
			Palaiseau, France \\
			lucas.schott@irt-systemx.fr}
		\and
		\IEEEauthorblockN{Hatem Hajri}
		\IEEEauthorblockA{\textit{
				Institute of Research } \\
			\textit{and Technology SystemX} \\
			Palaiseau, France \\
			hatem.hajri@irt-systemx.fr} \\
		\and
		\IEEEauthorblockN{ }
		\IEEEauthorblockA{ }
		\and
		\hspace{4.5cm}
		\IEEEauthorblockN{Sylvain Lamprier}
		\IEEEauthorblockA{\textit{
				\hspace{4.5cm} Sorbonne University} \\
			\hspace{4.5cm}
			Paris, France \\
			\hspace{4.5cm}
			sylvain.lamprier@lip6.fr}
		\and
		\hspace{-1.91cm}\IEEEauthorblockN{Patrick Gallinari}
		\IEEEauthorblockA{\textit{
				\hspace{-1.91cm}
				Sorbonne University} \\
			\hspace{-1.91cm}
			Paris, France \\
			\hspace{-1.91cm}
			patrick.gallinari@lip6.fr}
	}
	
	\maketitle
	
	\vspace{-5cm}
	
	\begin{abstract}
		This paper introduces stochastic sparse adversarial attacks (SSAA), standing as simple, fast and purely noise-based targeted and untargeted attacks of neural network classifiers (NNC). SSAA offer new examples of sparse (or $L_0$) attacks for which only few methods have been proposed previously. These attacks are devised by exploiting a small-time expansion idea widely used for Markov processes. Experiments on small and large datasets (CIFAR-10 and ImageNet) illustrate several advantages of SSAA in comparison with the-state-of-the-art methods. For instance, in the untargeted case, our method called Voting Folded Gaussian Attack (VFGA) scales efficiently to ImageNet and achieves a significantly lower $L_0$ score than SparseFool (up to $\frac{2}{5}$) while being faster. Moreover, VFGA achieves better $L_0$ scores on ImageNet than Sparse-RS when both attacks are fully successful on a large number of samples. 
	\end{abstract}
	
	
	\section{Introduction}
	
	
	
	Adversarial examples in machine learning have been essential in improving robustness of neural networks in recent years. Most of the work in this topic has been centered around three categories of attacks according to the minimised distance between original and adversarial samples: $L_2$ (squared error) \cite{7780651,CW2}, $L_{\infty}$ (max-norm) \cite{FGM,PGD,BIM} and much less $L_0$ (or sparse) attacks (minimising the number of modified components). For $L_0$ attacks, a list of the most influential works, also related to our paper might be given  \cite{CW2,JSMA,DBLP:conf/cvpr/ModasMF19,Bibi_2018_CVPR,2020arXiv200611776A,2020arXiv200706032C,Croce_2019_ICCV,croce2020sparsers,dong2020greedyfool}. \\

	
	For a NNC $F: \mathbb R^n \rightarrow \mathbb R^p$, the predicted label for an input $x$ is  $\text{label}(x)=\underset{k}{\text{argmax}} \, F_k(x)$, where $F_1,\cdots,F_p$ are the class probabilities of $F$. We recall that an adversarial example to $x$ is an item $x^*$ such that $\text{label}(x^*) \neq \text{label}(x)$ (untargeted attack), or such that $\text{label}(x^*)=c$, with $c\neq\text{label}(x)$ a specific class (targeted attack).\\
	
	Sparse alterations can be encountered in many situations and have been motivated in the previous works. For instance, they could correspond to some raindrops on traffic signs that are sufficient to fool an autonomous driver \cite{DBLP:conf/cvpr/ModasMF19}.  Understanding  these special perturbations is fundamental to mitigate their effects and take a step forward trusting neural networks in real-life.\\
	
	This paper presents a general probabilistic approach to generate new 
	$L_0$ attacks which rely on random noises. We argue that existing  deterministic attacks, which classically perform by sequentially applying  maximal perturbations on selected  components of the input, fail at reaching accurate adversarial examples on real-world large scale datasets. \\
	
	
	\begin{figure}[H]
		\centering
		\subfloat[Targeted XSMA (base-10 log scale)]  
		{\includegraphics[scale=0.24]{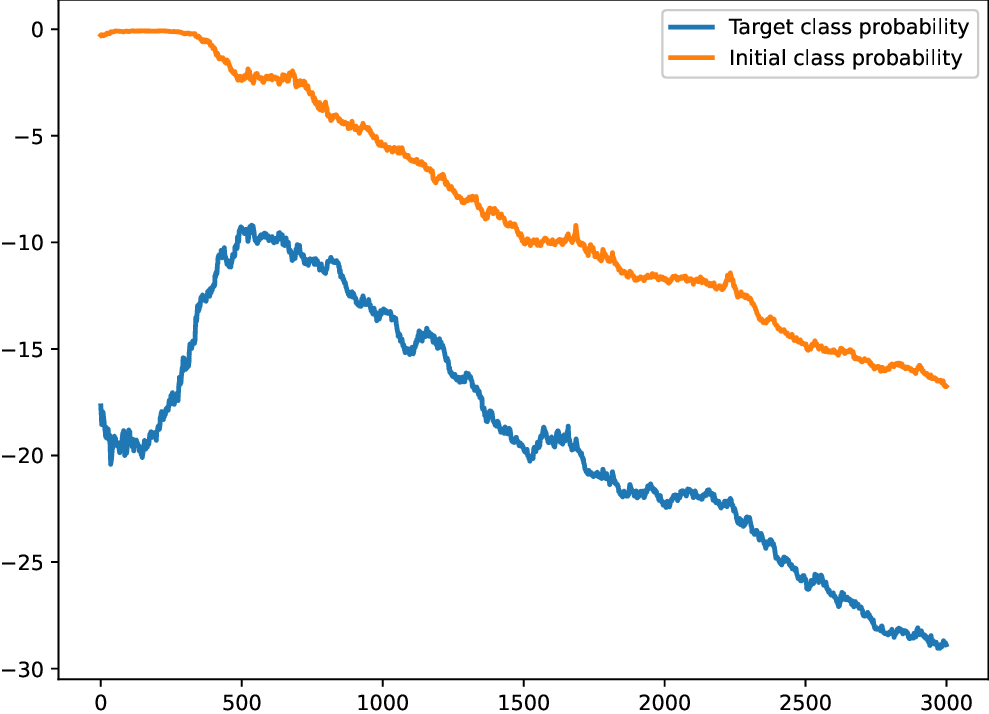}\label{VGG-p_worst}}
		\hspace{0.5cm}
		\subfloat[Targeted VFGA10]{\includegraphics[scale=0.24]{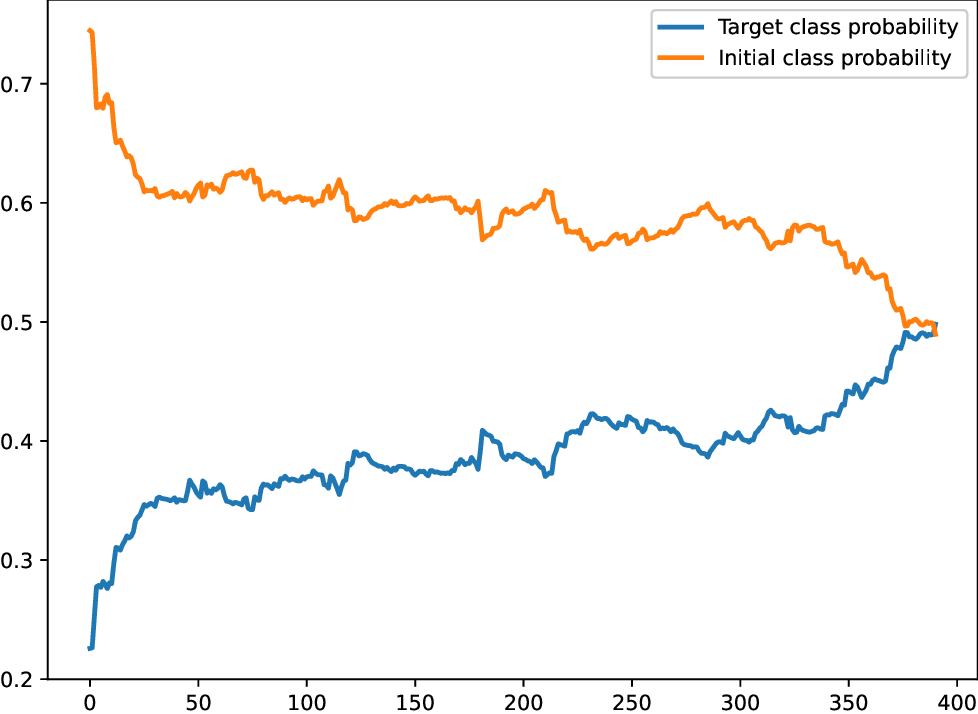}\label{VGG-_worst_pert}}
		\caption{Plots of the initial and targeted class probabilities for a one pixel version of XSMA on the left failing to converge along more than 3,000 iterations and our VFGA10 method converging efficiently in \textbf{less than 400 iterations} on the right.}
		\label{fig:long1}
	\end{figure}
	
	
	Figure \ref{fig:long1} (left) illustrates this failure on the ImageNet dataset \cite{ILSVRC15} for a  one-component version of the targeted XSMA attacks (JSMA \cite{JSMA}, WJSMA, TJSMA \cite{2020arXiv200706032C}) which does not succeed to affect the initial probability of the input on the Inception-v3 network \cite{DBLP:journals/corr/SzegedyVISW15}. On the other hand, working with more than one component at a time, while more accurate, does not scale at all on datasets as ImageNet. An alternative would be to repeatedly apply very small perturbations on components, but this would be at the cost of efficiency. Our claim is that random attacks, while not much studied in the literature of $L_0$ adversarial attacks, are able to cope with these issues. \\      
	
	
	Stochastic sparse adversarial attacks (SSAA) are inspired by the study of stochastic diffusions, their infinitesimal generators and boundary behaviors. They follow main existing $L_0$ attacks, which rely on iteratively selecting the most salient input feature by means of saliency maps, but consider probabilistic distributions for component and intensity selections. After identifying the best component to alterate first, the process samples intensities of perturbations for the selected component and chooses the best move among them. This allows to obtain accurate adversarial samples more efficiently than approaches based on deterministic perturbations. Experimental results on large scale datasets, as depicted on the same example as the failure case of XSMA in Figure \ref{fig:long1} (on the right), show that our SSAA approaches (denoted VFGA10) succeed at efficiently producing accurate attacks in most cases. \\ 
	
	
	The rest of the paper is organised as follows. Section \ref{sec2} introduces our SSAA. In Sections \ref{sec4} and \ref{sec40}, we experiment these attacks on deep NNC on CIFAR-10 \cite{cifar10} and ImageNet \cite{ILSVRC15} and compare their performances with the-state-of-the-art methods SparseFool \cite{DBLP:conf/cvpr/ModasMF19}, GreedyFool \cite{dong2020greedyfool}, Brendel \& Bethge $L_0$ attack (B$\&$B) \cite{NEURIPS2019_885fe656} and Sparse-RS \cite{croce2020sparsers}. Experimental results show that our untargeted VFGA scales efficiently to ImageNet and outperforms SparseFool while being faster. Furthermore, VFGA  achieves better $L_0$ scores on ImageNet than Sparse-RS when both attacks are fully successful on a large number of samples in the untargeted/targeted case. It is significantly less complex than B$\&$B and GreedyFool and obtains competitive results in some cases. Finally, Section \ref{sec7} presents a conclusion and possible continuations of this work.
	
	Our findings demonstrate that, unlike ongoing 
	works \cite{Bibi_2018_CVPR,2020arXiv200611776A} to introduce adversarial attacks at the level of the state-of-the-art while using random noises, our methods are able to reach and by-pass the state-of-the-art ones.\\
	
	\vspace{-0.2cm}
	
	\section{Stochastic sparse Attacks} 
	\label{sec2}
	
	In this section, we introduce SSAA by means of Gaussian noises on selected components of the input. To simplify the presentation, we mainly discuss targeted attacks and then deduce untargeted ones by applying slight modifications. The aim herein is to iteratively identify the best component to perturb and the best move for this component until the target label becomes the most probable for the NNC.\\
	
	Consider a Gaussian noise $X_\theta \sim \mathcal N(0,\theta)$ and denote by $(e_1,\cdots,e_n)$ the basis of $\mathbb R^n$. Any $c$-targeted probability expectation of the perturbed input $x+X_\theta e_i$ can be expanded as follows: 
	\begin{equation}\label{eq1000}
		\mathbb E[F_c(x+X_\theta e_i)]=F_c(x)+ \theta (\mathcal G_i F_c)(x) + ...
	\end{equation}
	\vskip 0.2cm
	\noindent
	where $\mathcal G_i F_c =  \dfrac{1}{2} \dfrac{\partial^2 F_c}{\partial x_i^2}$ is the infinitesimal generator of $X_\theta$ seen as a diffusion. When taking the folded Gaussian noise $X_\theta\sim|\mathcal N(0,\theta)|$, this expansion becomes:
	\begin{equation}\label{eq1001}
		\mathbb E[F_c(x+X_\theta e_i)]= F_c(x)+\sqrt{\frac{2\theta}{\pi}} \frac{\partial F_c}{\partial x_i}(x) + \theta (\mathcal G_i F_c)(x) + ...
	\end{equation}
	We build our reasoning upon a heuristic which is to look for the input feature $i$ that maximizes 
	$\mathbb E[F_c(x+X_{\theta} e_i)]$. The assumption behind this heuristic is that searching for the best expectation will allow to discover the best moves according to the distribution of the noise $X_\theta$.  This does not hold for the Gaussian noise $X_\theta \sim \mathcal N(0,\theta)$, since in that case the approximations $\mathbb E[F_c(x+X_\theta e_i)] \approx F_c(x)+ \theta (\mathcal G_i F_c)(x)$ and   
	$Var[F_c(x+X_\theta e_i)] \approx \theta \left(\dfrac{\partial F_c}{\partial x_i}\right)^2(x) $  are  of the same order as  $\theta$, indicating that variance should be taken into account in selecting the best components to perturb. On the other hand, considering the folded Gaussian noise $|X_\theta|$ and using the approximation $\mathbb E[F_c(x+X_\theta e_i)]\approx F_c(x)+\sqrt{\dfrac{2\theta}{\pi}} \dfrac{\partial F_c}{\partial x_i}(x)$ induces a negligible variance (only terms of $\theta^p$ with $p\ge 1$) in front of the expectation, at least  when $|\theta|<1$.  This means that working with the folded Gaussian distribution allows us to only focus on the expected probability of the perturbed input. Note also that the approximation of this expected probability only contains first derivatives w.r.t. to the input component which is a practical advantage of the folded over the pure Gaussian noise.\\
	
	While it would have been possible to consider some combination of  $\mathbb E[F_c(x+X_\theta e_i)]$ and $Var[F_c(x+X_\theta e_i)]$ for the Gaussian noise, taking a folded noise presents an important additional advantage for bounded inputs. Please note that, without loss of generality, we consider inputs  bounded in $[0,1]$ in this paper, as well as  the adversarial samples which share the same support domain. In the following, we propose to automatically tune the variance  parameter  $\theta$ of $X_\theta$ according to the distance of the input $x_i$ to these bounds.  Please note that, for a given component $i$, $x_i \neq 0.5 $, the possible amplitude of move is not the same in both directions. Considering a Gaussian noise, since symmetric,  would be problematic for this $\theta$ tuning. Rather, considering two folded Gaussian noises for each component, one positive (only for component increase) and one negative (only for component decrease) allows better fitted selections.\\
	
	In the following, we first present a one-sided, only increasing perturbations, stochastic attack based on folded Gaussian noises. Then, we deduce a both-sides attack, that considers the best choice between increase and decrease of each component, called Voting Folded Gaussian Attack.\\   
	
	\vspace{-0.2cm}
	
	\subsection{Folded Gaussian Attack (FGA)}

	
	For our one-side targeted attack FGA, the most relevant input feature to perturb  is thus selected by the rule $i=\underset{j}{\text{argmax}} \; \sqrt{\theta_j} \, \dfrac{\partial F_c}{\partial x_j}$, considering a folded Gaussian noise $|\mathcal N(0,\theta_{i})|$.\\

	\noindent\textbf{Choosing the variance $\theta_i$.}  Since FGA only considers positive perturbations of the input, fixing the variance $\theta_i$ must consider the upper-bound of the input domain. A quite natural choice could be either $\theta_i=1-x_i$ (variance = $1-x_i$) or $\sqrt{\theta_i}=1-x_i$ (standard deviation = $1-x_i$). We choose $\sqrt{\theta_i}=1-x_i$ to ensure that a generated perturbation $x_i+\mathcal N_i$ to $x_i$ has probability $2/3$ to be inside the interval $[x_i,1]$ (before clipping to $[0,1]$) which is a more motivated choice. Our experimental results (not reported in this paper) show that this choice gives slightly more effective attacks than the second one.\\
	
	\vspace{-0.2cm}
	After selecting the input feature $i$, our proposal is to simulate $N_S$ samples from  
	$|\mathcal N(0,\theta_i)|$  to find an accurate move towards a close adversarial sample. The complete process is depicted in Algorithm \ref{algo_levy} introducing the increasing FGA (and the decreasing FGA by analogy).  \\
	
	\begin{algorithm}[H]
		\DontPrintSemicolon
		
		\KwInput{$x$: input of label $l$, $c\ne l$: targeted class. \\ \hspace{0.8cm} $N_S$: number of samples to generate.\\ 
			\hspace{0.8cm} $\mathtt{maxIter}$: maximum number of iterations.}
		\KwOutput{$\tilde{x}$: adversarial sample to $x$.}
		\vspace{0.2cm}
		\KwInit{$\tilde{x}\leftarrow x,$ \;  $\Gamma\leftarrow\{1,\cdots,{\rm dim}(x)\}\setminus\{i : \tilde{x}_i=1\},$ \; $\mathtt{iter} \leftarrow 0$.}
		
		\While{$\Gamma\ne \emptyset,\, \text{label}(\tilde{x})\ne c$ and $\mathtt{iter}<\mathtt{maxIter}$}
		{
			$i_0=\underset{{i\in\Gamma}}{\text{argmax}} \; (1-\tilde{x}_i)\,\dfrac{\partial F_c}{\partial x_i}(\tilde{x})$.\;
			
			Generate samples $\left(S^h\right)_{1\le h \le N_S}$ from $|\mathcal N(0,\,\theta_{i_0})|$ \\ where  $\sqrt{\theta_{i_0}}:=1-\tilde{x}_{i_0}$.\;
			
			\vspace{0.3cm}
			\For{$h\in[\![1,\,N_S]\!]\vspace{0.2cm}$}
			{$\textrm{Define the input} \;\; \tilde{y}^{\,h} \;\; \textrm{by } \vspace{0.3cm}\\
				\left\{
				\begin{array}{ll}
					\tilde{y}^{\,h}_{j}\gets \text{Clip}_{[0,1]}\left(\tilde{x}_{j}+S^h\right) & \mbox{\bf if } j=i_0 \\
					[0.5em]\tilde{y}^{\,h}_j = \tilde{x}_j & \mbox{\bf otherwise.}
				\end{array}
				\right.$ \;
				
			}
			\vspace{0.1cm}
			Batch compute $F_c\left(\tilde{y}^{\,h} \,;\, h\in[\![1,\,N_S]\!]\right)$.\;
			\vspace{0.1cm}
			
			$\tilde{x}\leftarrow\underset{{\tilde{y}^{\,h}}}{\text{argmax}} \, F_c\left(\tilde{y}^{\,h} \right)$ , \;        	$\Gamma\leftarrow\Gamma\setminus\{i_0\}$ \\ 
			$\mathtt{iter} \leftarrow \mathtt{iter} + 1$. \;
			
		}
		{\bf return } $\tilde{x}$
		\caption{(Increasing) Folded Gaussian Attack (FGA)}
		\label{algo_levy}
	\end{algorithm}
	
	\vspace{0.1cm}
	
	\noindent\textbf{Choosing $N_S$.} The number $N_S$ is the main hyperparameter of Algorithm \ref{algo_levy}. 
	Given its definition, one can expect that increasing it will increase, up to saturation, the effectiveness of the attacks. This may, however, slow down their speeds. Thanks to batch computing, with sufficient memory, Step 8 can be performed at the cost of $N_S=1$ and (reasonably) augmenting $N_S$ can make Algorithm \ref{algo_levy} converge faster as less iterations would be needed. In most of our experiments, we fix this number to $N_S=10$ but also address some comparisons with $N_S=20, 100$. We refer to the analysis of the experimental results for more discussions related to this point. Finally, we also notice that batch computing used here does not often require a parallel computing effort by the user as this option is available in standard libraries.\\
	
	
	\vspace{-0.2cm}
	While the previous process only applies perturbations that increase the input, lowering the input features intensities can be as effective as increasing them. Following the same analogy, we introduce the decreasing FGA attack by taking $\sqrt{\theta_i}=x_i$ rather than $\sqrt{\theta_i}=1-x_i$ and replacing $|\mathcal N(0,\theta_i)|$ with  $-|\mathcal N(0,\theta_i)|$ in the previous algorithm. Note that FGA and XSMA are one sided attacks but, while XSMA apply predefined maximal perturbations, FGA explores in real time best perturbations to apply. \\

	\subsection{Voting Folded Gaussian Attack (VFGA)} 
	In this section, we propose a two-sided attack, which both considers  $\mathbb E[F_c(x+|X_{\theta_i^+}|\, e_i)]$ and $\mathbb E[F_c(x-|X_{\theta_i^-}| \,e_i)]$ for each feature, with $X_{\theta} \sim \mathcal N(0,\theta)$, $\sqrt{\theta_i^+}=1-x_i$ and $\sqrt{\theta_ i^-}=x_i$. This method applies increasing and decreasing FGA at each iteration and chooses the most effective moves in both directions. Details are given in Algorithm \ref{hhz}.\\
	
	
	
	\begin{algorithm}[H]
		\DontPrintSemicolon

		\KwInput{$x$: input of label $l$, $c\ne l$: targeted class.\\
			\hspace{0.8cm} $N_S$: number of samples to generate.\\ 
			\hspace{0.8cm} $\mathtt{maxIter}$: maximum number of iterations.}
		\KwOutput{$\tilde{x}$: adversarial sample to $x$.}
		\vspace{0.2cm}
		\KwInit{$\tilde{x}\leftarrow x,$ \;  $\Gamma\leftarrow\{1,\cdots,{\rm dim}(x)\}\setminus\{i : \tilde{x}_i=1\},$ \; $\mathtt{iter} \leftarrow 0$.}
		
		\While{$\Gamma\ne \emptyset,\, \text{label}(\tilde{x})\ne c$ and $\mathtt{iter}<\mathtt{maxIter}$}
		{
			$i^+=\underset{{i\in\Gamma}}{\text{argmax}}\, (1-\tilde{x}_i)\,\dfrac{\partial F_c}{\partial x_i}(\tilde{x}), \ \ i^-=\underset{{i\in\Gamma}}{\text{argmin}}\, \tilde{x}_i\,\dfrac{\partial F_c}{\partial x_i}(\tilde{x})$. \;
			
			\vspace{0.1cm}
			Generate samples $\left(S^{+,\,h}\right)_{1\le h \le N_S}$ from  $|\mathcal N(0,\,\theta_i^+)|$ where  $\sqrt{\theta_i^+}:=1-\tilde{x}_{i^+}$.\;
			
			Generate samples $\left(S^{-,\,h}\right)_{1\le h \le N_S}$ from $-\,|\mathcal N(0,\,\theta_i^-)|$ where  $\sqrt{\theta_i^-}:=\tilde{x}_{i^-}$. \;
			
			\vspace{0.3cm}
			\For{$h\in[\![1,\,N_S]\!]\vspace{0.2cm}$} 
			{$\textrm{Define the input} \;\; \tilde{y}^{\,h} \;\; \textrm{by } \vspace{0.3cm} \\ 
				\left\{
				\begin{array}{ll}
					\tilde{y}^{\,\pm,\, h}_{j}\gets \text{Clip}_{[0,1]}\left(\tilde{x}_{i^\pm}+S^{\pm,\,h}\right) & \mbox{\bf if } j = i^\pm \\
					[0.6em]\tilde{y}^{\,\pm,\, h}_{j} = \tilde{x}_{j} & \mbox{\bf otherwise.}
				\end{array}
				\right.$ \;
				
			}
			\vspace{0.1cm}
			Batch compute $F_c\left(\tilde{y}^{\,\pm,\,h} \,;\, h\in[\![1,\,N_S]\!]\right)$. \;
			
			\vspace{0.1cm}
			$\tilde{x}\leftarrow\underset{{\tilde{y}^{\,\pm,\,h}}}{\text{argmax}} \, F_c\left(\tilde{y}^{\,\pm,\,h} \right),$ \;
			$\Gamma\leftarrow\Gamma\setminus\{i_0\}$  with $i_0=i^+$ or $i^-$ according to the best move; $\mathtt{iter} \leftarrow \mathtt{iter} + 1$. \;
			
		}
		{\bf return } $\tilde{x}$
		\caption{Voting Folded Gaussian Attack (VFGA)}\label{hhz}
	\end{algorithm}

	\vspace{0.2cm}
	
	\subsection{Untargeted SSAA}\label{sec4} 
	The main focus for these attacks is to decrease the class probability of the input until a new class label is found. Few modifications are required to deduce the untargeted versions of the previous Algorithms: by assuming $c$ is the true label of $x$ and replacing $\text{argmax}$ with $\text{argmin}$ in Steps 3 and 9 of Algorithm \ref{algo_levy} and making similar slight changes in Algorithm \ref{hhz}.\\
	
	\section{Experiments on untargeted attacks}\label{sec4}
	In this section, we present experiments to highlight the benefits of our untargeted attacks. First, we aim to showcase the relevance of FGA in comparison with an alternative approach that uses the uniform noise called UA. Second, we aim to compare our attacks and more specifically VFGA with relevant state-of-the-art approaches. To this end, we will need to distinguish between two categories of methods: (1) fast and (2) more slow/complex methods. The code is available at \url{https://github.com/hhajri/stochastic-sparse-adv-attacks}. \\
	
	In the experiments, we consider two popular computer vision datasets illustrating small and high dimensional data: CIFAR-10 \cite{cifar10} (32 $\times$ 32$\times$ 3 images divided into $10$ classes) and ImageNet \cite{ILSVRC15} (ILSVRC2012 dataset containing 299 $\times$ 299$\times$ 3 images divided into 1,000 classes). The used neural network classifiers are described in the upcoming paragraphs.\\
	
	The state-of-the-art attacks considered for comparison in this section are: \\
	
	\noindent\textbf{SparseFool \cite{DBLP:conf/cvpr/ModasMF19}.}  This method is fast and scalable. At each iteration, it applies DeepFool \cite{7780651} to estimate the minimal adversarial perturbation thanks to a linearization of a classifier. Then, it estimates the boundary point and the normal vector of the decision boundary and finally updates the input features with a linear solver.\\
	
	\noindent\textbf{Brendel \& Bethge $L_0$ attack (B$\&$B) \cite{NEURIPS2019_885fe656}}. This gradient-based adversarial attack follows the boundary between the space of adversarial and non-adversarial images to find the minimum distance to the clean image. It is powerful and more efficient (but also slower and more complex) than many gradient-based approaches such as SparseFool. \\
	
	\noindent \textbf{GreedyFool \cite{dong2020greedyfool}.} This attack is an improvement of SparseFool. It is however more complex than the later as it needs to carefully train a distortion map which is a generative adversarial network GAN \cite{NIPS2014_5ca3e9b1}. We remark (based on one experiment on ImageNet) that it is less efficient (but also faster and less complex) than B$\&$B. \\
	
	\noindent\textbf{Sparse-RS \cite{croce2020sparsers}.} This attack is fast and achieves high success rate on ImageNet outperforming many white-box attacks such as $\text{PGD}_0$ \cite{Croce_2019_ICCV}. It requires fixing the maximum number of pixels to modify which is then fully exploited. In order to generate adversarial examples with minimal $L_0$ perturbations by Sparse-RS, one needs to run this method for several budgets before selecting a convenient one. \\

	\textbf{A notable difference with Sparse-R.} It should be mentioned that our attacks and Sparse-RS follow different strategies. Indeed, the budget $k$ for Sparse-RS is fixed in the pixel space. For instance, on CIFAR-10 this can go up to $32\times 32$, and once $k$ is fixed the number of modified pixels in the input space, for Sparse-RS, is near $3\times k$. Our attacks compute perturbations directly in the input space. All attacks are however $L_0$ in the usual definition and they are compared according to the most commonly used metric which is, up to our knowledge, the $L_0$ distance in the input space. 
	\\
	
	
	All the previous attacks are experimented using the original (PyTorch) implementations by the authors and following the recommended hyperparameters. Our attacks are also implemented with PyTorch. \\
	
	Finally, we introduce the attack based on the uniform noise:\\
	
	\noindent\textbf{Uniform attack (UA). } This method adds random uniform noises instead of folded Gaussian ones. It follows the lines of Algorithm \ref{algo_levy} (in its untargeted form), but Step 3 is replaced with $i_0=\underset{{i\in\Gamma}}{\text{argmin}} \, (1-x_i)\,\dfrac{\partial F_c}{\partial x_i}(x)$ and sampling in Step 4 is done from $\mathcal U([0,\theta_i]),\; \theta_i=1-x_i$.\\
	
	To compare between the different methods, we rely on the following scores: success rate (SR), mean/median number of changed pixels (${\rm Mean}$ and ${\rm Median}$), complexity based on the number of model propagation \cite{DBLP:conf/icml/Croce020a} (MP). We prefer MP over the running time per image since the later depends very much on the software used when executing the codes. \\
	
	\noindent\textbf{More approaches.} In this paper, since we propose fast methods, we only focus on comparisons with similar fast approaches like SparseFool and  Sparse-RS. The B$\&$B, although not fast, has been selected as a highly efficient benchmark attack. We omit comparison with Carlini$\&$Wagner $L_0$ \cite{CW2} and we believe the results would be similar to our comparison with B$\&$B. Also, we omit comparison with CornerSearch \cite{Croce_2019_ICCV} because it is less effective than Sparse-RS based on the work \cite{croce2020sparsers} and also needs a large computational cost on ImageNet (see results in \cite{Croce_2019_ICCV}).\\
	
	\subsection{On CIFAR-10.}
	\label{seccifar}
	On this dataset, we use the ResNet18 \cite{DBLP:conf/cvpr/HeZRS16} and VGG-19  \cite{DBLP:journals/corr/SimonyanZ14} models. After training with PyTorch, these networks reached $95.55 \%$ and {\bf $93.87 \%$} accuracies  respectively. For our attacks UA, FGA and VFGA, the hyperparameter $N_S$ is fixed to $N_S=10$ in the experiments (the obtained attacks are denoted UA10, FGA10 and VFGA10). The effect of augmenting $N_S$ is analysed later on in this section. We notice that, otherwise stated, $\mathtt{maxIter}$ is put to its maximal value in all the paper. The state-of-the-art approaches outlined before, except GreedyFool, are tested and compared with our methods on the correctly predicted samples among the $10,000$ CIFAR-10 test images. We refrained from comparing with GreedyFool because of the need to train the distortion map network on CIFAR-10 not provided in the code of \cite{dong2020greedyfool} (this network has been made available for the ImageNet dataset and comparisons on this dataset are considered in the next section). For Sparse-RS, several budgets of pixels $k$ (the number of pixels to modify) have been experimented and the smallest budget giving $100\%$ has been selected. On CIFAR-10, $k$ is optimal, i.e $k'=k-1$ does not give full success of the attack. Our intention is to show that under the condition of full success for all attacks (when possible) our VFGA method is overall more advantageous. \\
	
	
	\begin{table}[H]
		\centering
		\begin{tabular}{|M{2.5cm}|M{0.8cm}|M{1cm}|M{1.1cm}|M{1.1cm}|}
			\hline 
			Attacks &SR & ${\rm Mean}$ & ${\rm Median}$ & MP \\  
			\hline
			\hline
			\multicolumn{5}{|c|}{ ResNet18} \\
			\hline
			\textbf{B$\&$B} & 100 & 8.33 & 8.0 & 1927 \\
			\hdashline
			
			\textbf{SparseFool} & 99.31 & 36.48 & 9.0 & 520 \\
			\textbf{Sparse-RS ($k=10$)} & 100 & 29.79 & 30 & \textbf{GS} + 49 \\ 
			
			\hdashline
			\textbf{UA10} & 100 & 30.94 & 20.0 &  363 \\ 
			\textbf{FGA10}  & 100 & 29.70 & 20.0 &  134 \\ 
			\textbf{VFGA10}  & 100 & \textbf{17.03} & \textbf{11.0}  & \textbf{99} \\ 
			\hline
			\multicolumn{5}{|c|}{VGG-19} \\
			\hline
			\textbf{B$\&$B} & 100 & 5.30 & 6.0 & 1483 \\
			\hdashline
			
			\textbf{SparseFool}  & 97.98 & 67.71 & 8.0 & 686 \\
			\textbf{Sparse-RS ($k=7$)} & 100 & 20.82 & 21 & \textbf{GS} + 55 \\ 
			\hdashline
			
			\textbf{UA10} & 100 & 22.16 & 11.0 & 281 \\ 
			\textbf{FGA10}  & 100 & 19.67 & 11.0 & 103 \\ 
			\textbf{VFGA10} & 100 & \textbf{11.40} & \textbf{7.0} & \textbf{80} \\ 
			\hline
		\end{tabular}
		\vskip 0.3cm
		\caption{Results on the correctly predicted among the $10,000$ test images of CIFAR-10. SR is the success rate of the attack, ${\rm Mean}$, ${\rm Median}$ are the average and median number of modified pixels on successful samples and MP is the number of model propagations. \textbf{GS} is a greed-search to find optimal values of $k$ giving full success that took \textbf{several hours}. The highlighted results of our VFGA in comparison with the fast methods SparseFool and Sparse-RS are in bold.}
		\label{table:rescifar100}
	\end{table}
	
	
	\noindent\textbf{Comments.} 
	The previous results show that the folded Gaussian noise is more advantageous in attacking than the uniform noise and that combining two folded distributions is useful not only for the SR, ${\rm Mean}$ and ${\rm Median}$ but also for the model propagation score. Concerning the state-of-the-art methods: VFGA has less advantageous ${\rm Mean}$ and ${\rm Median}$ than B$\&$B (near 2 times greater for ${\rm Mean}$ and the gap is reduced for ${\rm Median}$). Nevertheless, it is up to $\frac{1}{20}$ less complex based on the MP score. Second, all of our methods and more particularly VFGA significantly outperform SparseFool. Regarding the comparison with Sparse-RS, we remark that VFGA has notable ${\rm Mean}$ and ${\rm Median}$ advantages (up to 3 times fewer) and is also less complex given the number of experiments carried for Sparse-RS to achieve full success (with minimal ${\rm Mean}$ and ${\rm Median}$). Notice also the difficulty to find a good $k$ with full success for Sparse-RS as the optimal value depends on each sample and high values impact the overall performance of this attack. An advantage of our attack is that this parameter is set automatically and is optimal for each sample.\\
	
	\noindent\textbf{A comparison between Sparse-RS and VFGA for different distortions.} As stressed before, we only focus on performances under the condition of full success which is usually reported to summarise the contribution of new methods. If we relax this condition, we remark that for small budgets $k$ when both VFGA and Sparse-RS are not fully successful, Sparse-RS outperforms VFGA in SR but VFGA obtains better ${\rm Mean}$ and ${\rm Median}$ which are always near $k$ for Sparse-RS. Starting from a $k$ which approaches full success, VFGA becomes more advantageous in SR, ${\rm Mean}$ and ${\rm Median}$. \\
	
	
	
	
	\begin{figure}[H]
		\centering
		\includegraphics[scale=0.27]{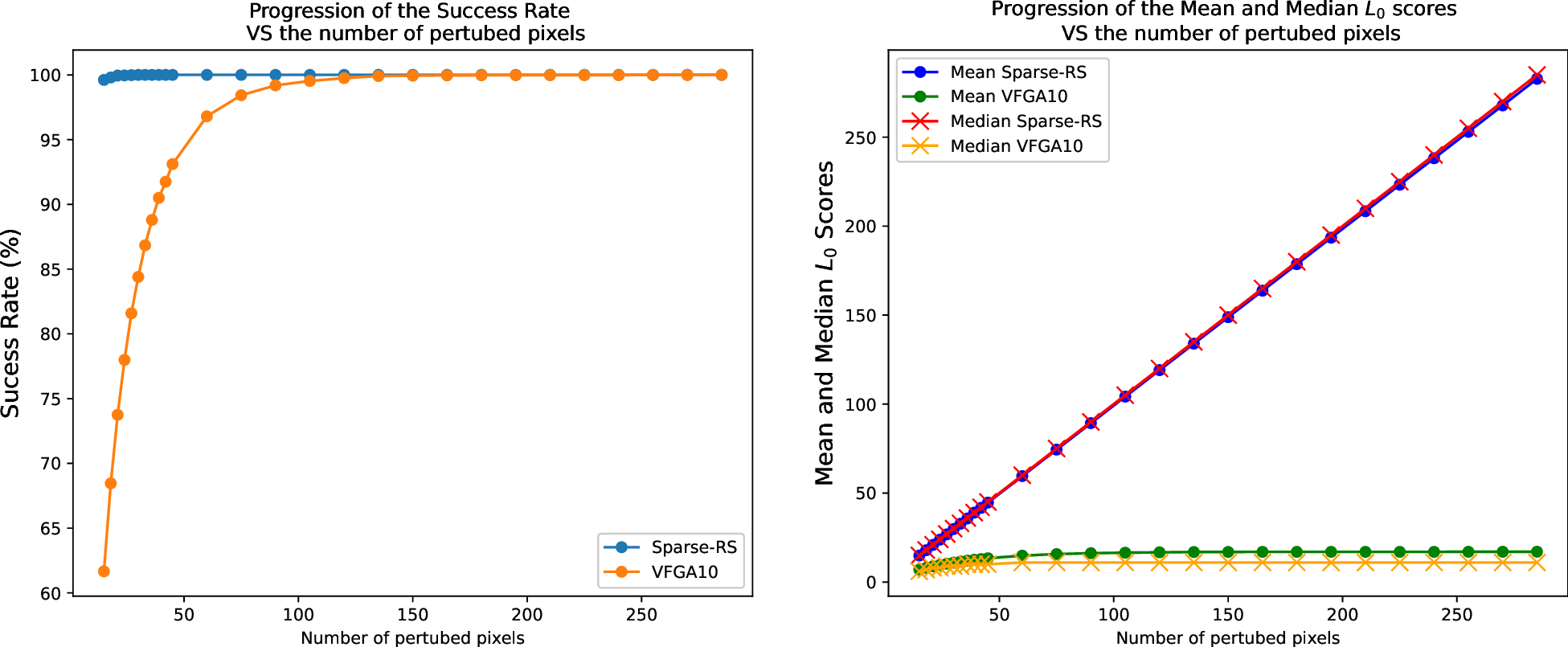}
		\caption{Results over the number of perturbed pixels for untargeted attacks ({\bf VFGA10 and Sparse-RS}) on CIFAR-10 for the ResNet-18 model. On the left, the success rate SR and on the right, the ${\rm Mean}$ and ${\rm Median}$ $L_0$ scores.}
		\label{VFGA_SparseRS}
	\end{figure}
	
	
	We draw in Figure \ref{VFGA_SparseRS} the SR (on the left), and ${\rm Mean}$ and ${\rm Median}$ $L_0$ scores (on the right) versus the number of perturbed pixels for VFGA10 and Sparse-RS. When only few pixels are perturbed, Sparse-RS performs better than VFGA10 in SR. Once the number of perturbed pixels exceeds $105$, the two SR become very competitive (near and then equal to $100\%$). The ${\rm Mean}$ and ${\rm Median}$ $L_0$ scores for VFGA10 are, however, always better than those by Sparse-RS regardless the number of pixels which have been modified.\\

	For Sparse-RS, ${\rm Median}$ is linear as a function of the number of perturbed pixels. For the budget of $285$ modified pixels, ${\rm Mean}$ is $282.78$. For VFGA10, this budget is the worst $L_0$ score among the 10,000 test images of CIFAR-10 and with this value, ${\rm Mean}$ is $17.03$ which is about $16$ times fewer than Sparse-RS. In short, even if for a small number of disturbed pixels, VFGA10 is not able to reach $100 \%$ in SR, its ${\rm Mean}$ and ${\rm Median}$ are still better than those of Sparse-RS. \\
	
	
	\noindent\textbf{Augmenting $N_S$.}	In what follows, we investigate the impact of augmenting $N_S$ on the performances of our attacks by testing UA20, FGA20 and VFGA20, which correspond to $N_S=20$, on the same data as Table \ref{table:rescifar100}. \\
	
	
	\begin{table}[H]
		\centering
		\begin{tabular}{|M{1.6cm}|M{0.8cm}|M{1cm}|M{1.3cm}|M{1.4cm}|}
			\hline 
			Attacks &SR & ${\rm Mean}$ & ${\rm Median}$ & MP. \\  
			\hline
			\hline
			\multicolumn{5}{|c|}{ ResNet18} \\
			\hline
			\textbf{UA20} & 100 & 30.93 & 20.0   & 684 \\ 
			\textbf{FGA20}   & 100 & 30.23 & 19.0  & 193 \\ 
			\textbf{VFGA20}   & 100 & 16.76 & 11.0 & 131 \\ 
			\hline
			\multicolumn{5}{|c|}{VGG-19} \\
			\hline
			\textbf{UA20} & 100 & 20.60 & 10.0  & 457 \\ 
			\textbf{FGA20}  & 100 & 19.71 & 10.0  & 134 \\ 
			\textbf{VFGA20}  & 100 & 11.22 & 7.0 & 113 \\ 
			\hline
		\end{tabular}
		\vspace{0.1cm}
		\caption{Comparison between our attacks for $N_S=20$.}
		\label{table:ns20cifar}
	\end{table}
	
	\vspace{-0.2cm}
	
	We remark that SR, ${\rm Mean}$ and ${\rm Median}$ are slightly improved but based on MP score, the attacks become more complex. This illustrates the fact that increasing so much $N_S$ may not significantly improve the attacks but on the other hand it may slow down them. Also, we observe that despite augmenting $N_S$ from $10$ to $20$, the uniform attack cannot beat FGA10. This is quite remarkable since for the uniform distribution the $N_S$ generated samples are different and fall inside the domain of the input features while for the folded Gaussian distribution, due to clipping, several samples are likely to be clipped at the minimal and maximal bounds. Augmenting $N_S$ also increases more quickly MP for the uniform noise. \\
	
	\subsection{On ImageNet.}
	\label{secimagenet}
	In this section, we test the ability of the previous attacks and additionally GreedyFool to generate adversarial examples at large scale by considering models on ImageNet. Two pre-trained networks provided by PyTorch are considered for testing: Inception-v3 \cite{DBLP:journals/corr/SzegedyVISW15} and VGG-16 \cite{DBLP:journals/corr/SimonyanZ14} whose accuracies are respectively $77.45 \%$ and $71.59 \%$. Inputs are of size 299$\times$299$\times$3 and 224$\times$224$\times$3 for the first and second model respectively. \\ 
	
	Again, we consider B$\&$B as a benchmark of a highly successful attack. GreedyFool requires training a GAN network on ImageNet but once carefully done it is highly successful. We recover the GAN model from the code of \cite{dong2020greedyfool} and complement the code to compute MP for this attack. For Sparse-RS, we again fix our objective to compare with this attack when 100$\%$ SR is achieved. This requires launching several experiments for different values of $k$ on the whole considered set of images in order to obtain a \textbf{near-optimal} value. By this, we mean a value $k$ giving $100\%$ SR ; there exists $k'<k$ and the performances of VFGA in ${\rm Mean}$ and ${\rm Median}$ are better than those obtained by Sparse-RS with budget $k'$. This implies in particular that VFGA gives better results than Sparse-RS when tested with the optimal value of $k$. To get an idea of the difference between our results and those by Sparse-RS, we always report the results for $k'$ and $k$ by Sparse-RS (in this section and next one). \\
	
	
	The obtained results for the different attacks are reported in Table \ref{table:resImageNet} and commented after.
	
	\newpage
	
	
	\begin{table}[H]
		\centering
		
		\begin{tabular}{|M{2.6cm}|M{0.7cm}|M{0.8cm}|M{0.9cm}|M{1.5cm}|M{1.7cm}|M{0.9cm}|M{1cm}|M{4cm}|}
			\hline 
			Attacks &SR & ${\rm Mean}$ & ${\rm Median}$ & MP. \\  
			\hline
			\hline
			\multicolumn{6}{|c|}{ Inception-v3} \\
			\hline
			\textbf{B$\&$B} & 100 & 43.96 & 37.0 & 5602\\ 
			\hdashline
			\textbf{GreedyFool} & 100 & 86.09 & 79.0 & \textbf{GAN} + 617 \\
			
			\textbf{SparseFool} & 100 & 348.16 & 167.5 & 2531\\
			
			\textbf{Sparse-RS ($k'=90$)} & 99.62 & 267.13 & 270.0 & \textbf{GS} + 341 \\
			\textbf{Sparse-RS ($k=100$) } & 100 & 297.12 & 300.0 & \textbf{GS} + 358 \\

			\hdashline
			\textbf{UA10}  & 100 & 335.19 & 101.0 & 3042 \\
			\textbf{FGA10}  & 100 & 323.27 & 102.0 &  744 \\
			\textbf{VFGA10}  & 100 & \textbf{198.25} & \textbf{64.0}  & \textbf{1133} \\
			\hline
			\multicolumn{5}{|c|}{VGG-16} \\
			\hline
			\textbf{B$\&$B} & 100 & 39.24 & 25.0  & 3416 \\
			\hdashline
			\textbf{GreedyFool} & 100 & 66.18 & 31.0 & \textbf{GAN} + 589 \\

			\textbf{SparseFool} & 100 & 216.21 & 164.0 & 1460 \\ 
			\textbf{Sparse-RS ($k'=60$)} & 99.78 & 179.01 & 180.0 & \textbf{GS} + 240 \\
			\textbf{Sparse-RS ($k=70$)} & 100 & 204.59 & 210.0 & \textbf{GS} + 246 \\

			\hdashline
			\textbf{UA10} & 100 & 150.04 & 85.0  & 2122 \\ 
			\textbf{FGA10}  & 100 & 140.15 & 82.0  & 986 \\ 
			\textbf{VFGA10}  & 100  & \textbf{77.85} & \textbf{43.0} & \textbf{709} \\ 
			\hline
		\end{tabular}
		\caption{Results on the firstly $6,000$ correctly predicted validation images of ImageNet. \textbf{GS} is a greed-search to near-optimal values of $k$ that took \textbf{several days}. \textbf{GAN} is a generative network trained on ImageNet. Our results in comparison with the fast methods SparseFool and Sparse-RS when fully successful are highlighted in bold. }\label{table:resImageNet}
	\end{table}
	
	
	\noindent\textbf{Comments.} First, B$\&$B achieves the best SR, ${\rm Mean}$ and ${\rm Median}$ scores. GreedyFool comes after but with the cost of training a GAN model on ImageNet. The complexity comparison between these two attacks is difficult to address and we only claim that both methods are significantly more complex than our approach (GreedyFool is complex to reproduce on new datasets). Despite this fact, we observe that on VGG-16 our VFGA has a gap of ${\rm Mean}$ and ${\rm Median}$ of less than 12 pixels which is relatively small. Among the methods shown in the previous table, our attacks and SparseFool are the fastest ones under the full success condition and when minimising at the same time ${\rm Mean}$ and ${\rm Median}$. Our attacks, obtain, however, overall better performances than SparseFool according to all metrics. Specifically, VFGA significantly outperforms SparseFool with respect to all scores. Moreover, despite the fact that we select  near-optimal values of $k$ for Sparse-RS, VFGA is still more advantageous regarding ${\rm Mean}$ and ${\rm Median}$ and also faster if the complexity of finding $k$ is added. Finally, we notice that after finding a good $k$ giving full success for Sparse-RS, this attack can not generate relevant adversarial examples with minimal $L_0$ distance $1$, while due to the flexibility of our attack, several such examples can be generated. This is a further advantage of our attack.\\ 
	
	\vskip 0.6cm
	

	
	\section{Experiments on targeted attacks}\label{sec40}
	Targeted attacks are more challenging than untargeted ones. The objective of this section is to compare (targeted) VFGA, our selected method, with (targeted) Sparse-RS as a fast attack outperforming several state-of-the-art methods \cite{croce2020sparsers}. We recall that SparseFool is not efficient as a targeted attack. We do not report results by FGA and UA but claim that FGA is still more relevant than UA and only omit to address a similar comparison as before. We do not report results by B$\&$B and GreedyFool as targeted attacks because of the need of the distortion map on CIFAR-10 for GreedyFool, the non ability to reproduce B$\&$B in the targeted mode and moreover since, we consider that these approaches are complex to reproduce on new datasets. Thus, we only focus on the comparison with Sparse-RS and defend our approach as an efficient fast method. Our main conclusion in this paragraph is that, for ImageNet which is more challenging, VFGA is still more relevant than Sparse-RS regarding the same previous metrics when both attacks are fully successful and despite the fact that a near-optimal value of $k$ is selected. On CIFAR-10, we conclude that Sparse-RS is more advantageous in ${\rm Mean}$ and ${\rm Median}$.\\
	
	We consider the same datasets and network models as before. To simplify the experiment,  we do not consider all possible target labels but, for each test dataset, we generate a list of random labels which were fixed once for all. Each input image is then attacked to have one desired label. For Sparse-RS, we again select a near-optimal $k$ in both experiments. This task took several hours on CIFAR-10 and several days on ImageNet. We report in Tables \ref{table:rescifar90} and \ref{table:250} the results obtained on CIFAR-10 and ImageNet. In Table \ref{table:rescifar90} VFGA100 is VFGA with $N_S=100$.\\
	
	
	\begin{table}[H]
		\centering
		\begin{tabular}{|M{2.5cm}|M{0.6cm}|M{1cm}|M{0.8cm}|M{1.6cm}|M{1.8cm}|M{0.8cm}|M{0.8cm}|M{1.2cm}|M{1.4cm}|}
			\hline 
			Attacks &SR & ${\rm Mean}$ & ${\rm Median}$ & MP \\  
			\hline
			\hline
			\multicolumn{5}{|c|}{ ResNet18} \\
			\hline
			
			\textbf{Sparse-RS ($k=30$)} & 100 & \textbf{89.43} & \textbf{90.0} & \textbf{GS} + 1678 \\
			
			\textbf{VFGA10}  & 100 & 641.49 & 174.0 & \textbf{13427} \\
			\textbf{VFGA100} & 100 & 154.43 & 105.0 & \textbf{20619} \\
			\hline
			\multicolumn{5}{|c|}{VGG-19} \\
			\hline
			\textbf{Sparse-RS ($k=25$)} & 100 &\textbf{ 73.17} & \textbf{75.0} & \textbf{GS} + 1123  \\
			
			\textbf{VFGA10} & 100 & 551.27 & 150.0  & \textbf{12213} \\
			
			\textbf{VFGA100} & 100 & 174.93 & 97.0 & \textbf{21417} \\
			\hline
		\end{tabular}
		\vskip 0.3cm
		\caption{Results on the correctly-predicted test images of CIFAR-10. \textbf{GS} is a greed-search to find near-optimal values of $k$ that took \textbf{several hours}. The results for $k'$ are not reported since Sparse-RS is here more advantageous in full success.}
		\label{table:rescifar90}
	\end{table}
	
	
	First, we notice that Sparse-RS obtains better ${\rm Mean}$ and ${\rm Median}$ than VFGA10. Increasing $N_S$ from $10$ to $100$ improves considerably VFGA but our results are still less better than Sparse-RS. Given the time needed to find the near-optimal values $k$, we claim that our attacks are still overall much faster than Sparse-RS to obtain full success with optimal ${\rm Mean}$ and ${\rm Median}$.\\

	\begin{table}[H]
		\centering
		
		\begin{tabular}{|M{2.8cm}|M{0.7cm}|M{1cm}|M{0.8cm}|M{1.3cm}|}
			\hline 
			Attacks &SR & ${\rm Mean}$ & ${\rm Median}$  & MP \\  
			\hline
			\hline
			\multicolumn{5}{|c|}{ Inception-v3} \\
			\hline
			\textbf{Sparse-RS ($k'=950$)} & 99.87 & 2671.19 & 2850.0 & \textbf{GS} + 6591 \\
			\textbf{Sparse-RS ($k=1000$)} & 100 & 2898.72 & 3000.0 & \textbf{GS} + 6899 \\
			\textbf{VFGA10}  & 100 & \textbf{2148.53} & \textbf{1843.45} & \textbf{21616} \\
			\hline
			\multicolumn{5}{|c|}{VGG-16} \\
			\hline
			\textbf{Sparse-RS ($k'=450$)} & 99.91 & 1278.45 & 1350.0 & \textbf{GS} + 6963 \\
			\textbf{Sparse-RS ($k=500$)} & 100 & \textbf{1398.56} & 1500.0  & \textbf{GS} + 7003 \\
			\textbf{VFGA10}  & 100 & 1436.02 & \textbf{1057.38} & \textbf{14223} \\
			\hline
		\end{tabular}
		\vspace{0.3cm}
		\caption{Results obtained on the 5,000 firstly correctly-predicted validation images of ImageNet. \textbf{GS} is a greed-search to find near-optimal values of $k$ that took \textbf{several days}.}  
		\label{table:250}
	\end{table}
	
	
	Our interpretation of Table \ref{table:250} is overall similar to Table \ref{table:resImageNet}. For Inception-v3, which is more challenging, VFGA10 outperforms Sparse-RS regarding all scores. On VGG-16 Sparse-RS only takes a slight advantage of ${\rm Mean}$. As for untargeted attacks, a notable advantage of our methods is the flexibility of our budget of modifiable pixels which allows us to generate adversarial examples with minimum $L_0$ distance while being $100\%$ successful on all samples.\\ 
	
	
	
	
	\section{Conclusion}\label{sec7}
	This paper introduced noise-based attacks to generate sparse adversarial samples to inputs of deep neural network classifiers. A first advantage of our methods is that they work as both untargeted and targeted attacks. Moreover, they are very simple to put in place and require fixing only one parameter whose interpretation is intuitive (the bigger the best up to saturation in performance). Our attacks are faster to apply on new models and datasets than existing approaches (SparseFool, GreedyFool) while assuring full success. They are much less complex than the-state-of-the-art method B$\&$B relying on the model propagation score (near $\frac{1}{20}$ on CIFAR-10 and $\frac{1}{5}$ on ImageNet) and achieve competitive results in some cases. Finally, in comparison with Sparse-RS, our attacks are flexible allowing to find an optimal budget of pixels for each input image and achieve full success with minimal $L_0$ scores.  \\

	Our methodology relies on a simple expansion idea that provides a close link between adversarial examples and Markov processes.  We believe it can be pursued in several ways. For instance, continuing with $L_0$ attacks, it could be interesting to explore other types of noises  such as Poisson or compound Poisson noises and study their relevance in the setting of adversarial examples. Combining different noises attacks in a voting way, although simple, can lead to powerful $L_0$ attacks. We leave these questions to possible future works. \\
	
	\noindent\textbf{Acknowledgements.} We are very grateful to Maksym Andriushchenko and Francesco Croce for useful comments and references. We would like to thank Th\'eo Combey for his help in Tensorflow simulations of the attacks.
	\bibliography{biblio1}
	\bibliographystyle{plain}

\end{document}